%% file: main.tex
\documentclass[10pt,twocolumn,letterpaper]{article}

\usepackage{iccv}
\usepackage{times}
\usepackage{epsfig}
\usepackage{graphicx}
\usepackage{amsmath}
\usepackage{amssymb}

\usepackage{algorithm}
\usepackage{algorithmic}
\usepackage{mathtools}
\usepackage{comment}
\usepackage[table,xcdraw]{xcolor}
\usepackage{booktabs}
\usepackage{graphics}
\usepackage{etoc}
\usepackage{pdfpages}
\usepackage{etoolbox}
\usepackage[nocompress]{cite}

\input{custom}

\usepackage[breaklinks=true,bookmarks=false]{hyperref}

\iccvfinalcopy 

\def\iccvPaperID{5009} 

\def\ourAbbrname{\textbf{MuRAL}}
\def\ourFullname{\textbf{Mu}lti-scale \textbf{R}egion-based \textbf{A}ctive \textbf{L}earning}

\ificcvfinal\fi

\begin{document}

\title{MuRAL: Multi-Scale Region-based Active Learning for Object Detection}


\author{
Yi-Syuan Liou$^1$ \qquad Tsung-Han Wu$^1$ \qquad Jia-Fong Yeh$^1$ \qquad Wen-Chin Chen$^1$ \qquad Winston H. Hsu$^{1,2}$
\\
\\
$^1$National Taiwan University \qquad $^2$Mobile Drive Technology
}

\maketitle

\begin{abstract}
Obtaining large-scale labeled object detection dataset can be costly and time-consuming, as it involves annotating images with bounding boxes and class labels. Thus, some specialized active learning methods have been proposed to reduce the cost by selecting either coarse-grained samples or fine-grained instances from unlabeled data for labeling. However, the former approaches suffer from redundant labeling, while the latter methods generally lead to training instability and sampling bias. To address these challenges, we propose a novel approach called Multi-scale Region-based Active Learning (MuRAL) for object detection. MuRAL identifies informative regions of various scales to reduce annotation costs for well-learned objects and improve training performance. The informative region score is designed to consider both the predicted confidence of instances and the distribution of each object category, enabling our method to focus more on difficult-to-detect classes. Moreover, MuRAL employs a scale-aware selection strategy that ensures diverse regions are selected from different scales for labeling and downstream finetuning, which enhances training stability. Our proposed method surpasses all existing coarse-grained and fine-grained baselines on Cityscapes and MS COCO datasets, and demonstrates significant improvement in difficult category performance.
\end{abstract}

\section{Introduction}
\label{sec:intro}

Object detection \cite{ren2015faster, redmon2017yolo9000, lin2017focal, liu2016ssd} is a fundamental computer vision task that involves locating and categorizing objects in images. However, annotating object detection data can be costly, as it requires drawing bounding boxes around objects and labeling each box with its corresponding class. To address these challenges, researchers have proposed various label-efficient solutions, including weakly supervised learning \cite{bilen2016weakly, ash2019deep, bilen2015weakly, song2014weakly, li2016weakly}, semi-supervised learning \cite{tang2021proposal, jeong2019consistency, sohn2020simple, li2020improving}, and active learning \cite{haussmann2020scalable, yuan2021multiple, wu2022entropy, kao2018localization, choi2021active, yu2022consistency, desai2020towards, laielli2021region}. Active learning, in particular, is a promising technique that actively acquires limited labeled data for downstream training for several rounds. By iteratively prioritizing data points that contribute most to improving model accuracy for labeling, active learning can achieve accurate models while reducing labeling costs.

\begin{figure}
    \centering
    \includegraphics[width=\linewidth]{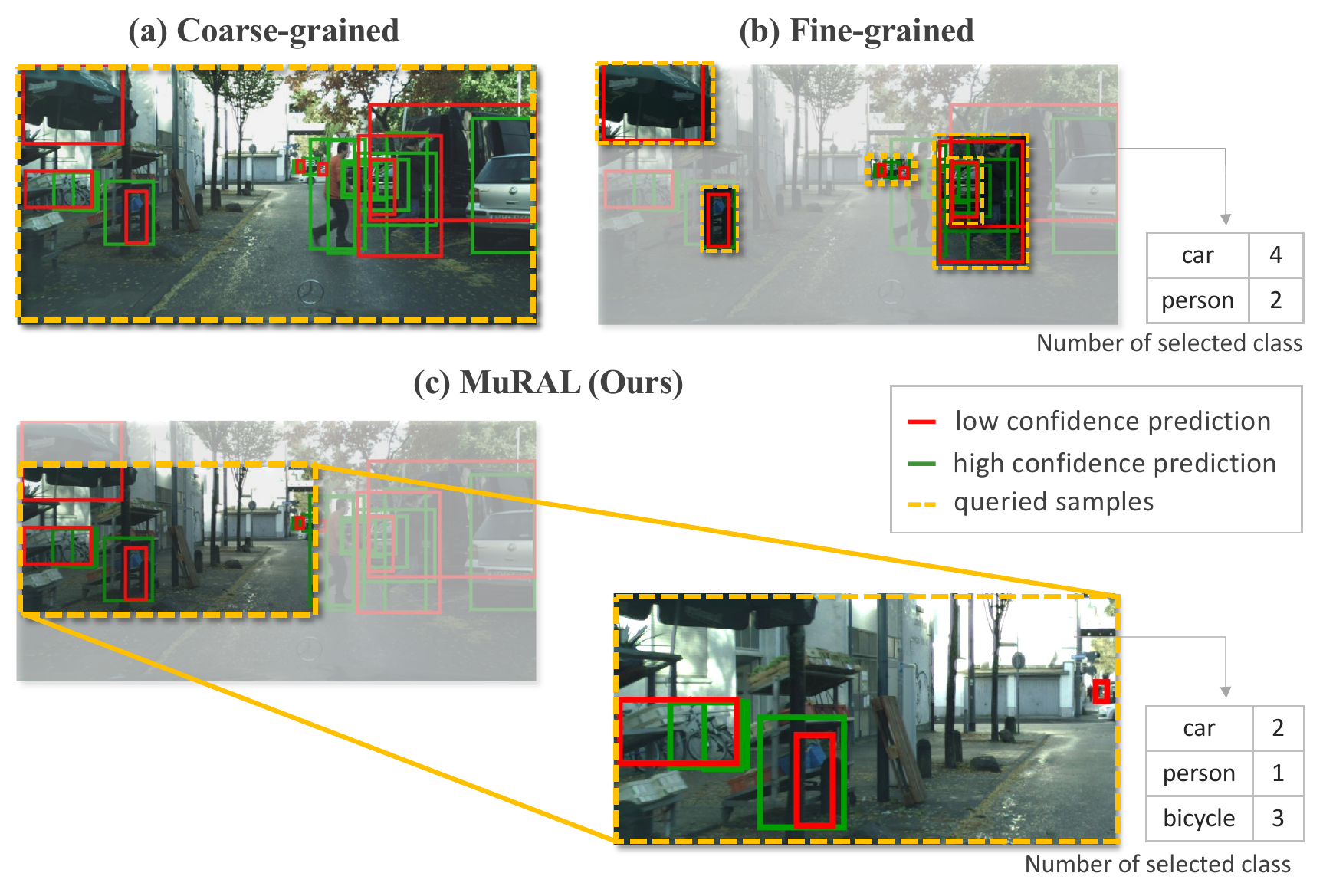}
    \caption{\textbf{Comparison to different active learning selection strategies.} 
(a) Selecting informative images for annotation using coarse-grained methods can result in redundant annotations of well-recognized objects within an image.
(b) Choosing bounding boxes to annotate using fine-grained methods often leads to sampling bias, where similar objects are acquired in a label query batch, and training instability due to incomplete labeling.
(c) Our \ourAbbrname{} integrates multi-scale regions into training to enhance stability and minimize the need for annotations on well-recognized objects. Furthermore, we select diverse regions categorically and spatially for labeling to prevent sample bias.}
    \label{fig:intro}
\end{figure}


In the context of object detection, current active learning methods can be classified into two main categories: coarse-grained and fine-grained approaches. Coarse-grained methods \cite{haussmann2020scalable, yuan2021multiple, wu2022entropy, kao2018localization, choi2021active, yu2022consistency} pick informative images for label acquisition based on classification or localization uncertainty, as illustrated in Fig. \ref{fig:intro} (a). However, this strategy may result in redundant labeling on well-detected instances. On the other hand, as depicted in Fig. \ref{fig:intro} (b), fine-grained methods \cite{choi2021active, yu2022consistency, laielli2021region} acquire bounding boxes annotations for specific instances or regions within an image and employ partially labeled images for further fine-tuning. Nevertheless, this approach may result in sampling bias and training instability due to partial labeling on certain classes or scales.

To address these challenges, we introduce a novel active learning framework: \ourFullname{} (\ourAbbrname{}) for object detection to overcome the limitations of coarse-grained and fine-grained methods. Unlike prior coarse-grained approaches \cite{haussmann2020scalable, yuan2021multiple, wu2022entropy, kao2018localization, choi2021active, yu2022consistency}, \ourAbbrname{} is able to effectively identify informative regions in an image, reducing annotation cost for well-learned objects. Compared to existing fine-grained strategies \cite{choi2021active, yu2022consistency, laielli2021region}, our framework picks categorically and spatially diverse regions for labeling to avoid sampling biases. In addition, as shown in Fig. \ref{fig:intro} (c), our proposed method employs multi-scale regions as queried samples to prevent incomplete label training issue and improve object detection performance for objects of various scales.

To elaborate, our active data selection strategy involves several steps, including multi-scale region candidate generation, informative score calculation, and scale-aware region selection. To begin, we consider region candidates as the fundamental units for labeling. The region candidates are generated at multiple scales across the image to highlight objects of varying sizes. Next, we rank these candidates within each scale based on their informative scores, which take into account the predicted confidence of bounding boxes covered in regions and the distribution of each object class. This enables us to identify and prioritize regions that are more informative and may contain difficult-to-detect object classes. Finally, we employ a scale-aware selection strategy to ensure that we query diverse regions of various sizes, promoting a stable training process and further enhancing the diversity of the sample set.

Extensive experiments demonstrate that our proposed \ourAbbrname{} outperforms existing coarse-grained and fine-grained active learning baseline methods and achieves state-of-the-art results on the Cityscapes \cite{cordts2016cityscapes} and MS COCO \cite{lin2014microsoft} object detection tasks. Moreover, ablation studies and additional analyses further highlight the effectiveness of all the components, particularly in selecting informative and diverse regions and improving the detection performance for difficult-to-detect categories in our proposed method. 

To sum up, the contributions of this paper are as follows,
\begin{itemize}
    \setlength\itemsep{0em}
    \item Our proposed multi-scale region-based active learning approach for object detection effectively tackles the issue of label redundancy in coarse-grained methods, as well as the problem of sampling bias and training instability in fine-grained methods.
    \item We employ a re-weighting class informative region score along with a scale-aware selection strategy to query informative and diverse multi-scale regions.
    \item Experimental results indicate that our method outperforms existing coarse-grained and fine-grained active learning methods on two datasets.
\end{itemize}

\begin{figure*}
\begin{center}
\includegraphics[width=\linewidth]{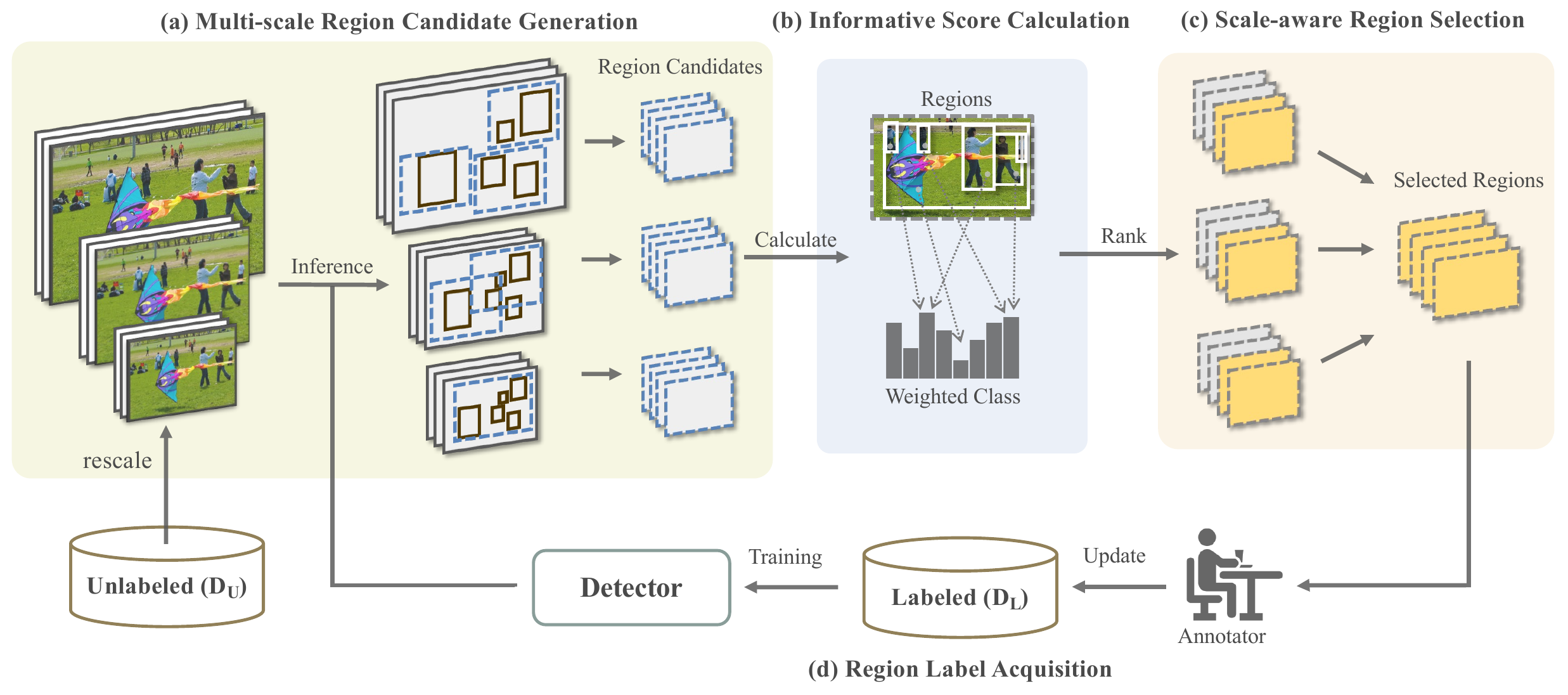}
\end{center}
   \vspace*{-3mm}
   \caption{\textbf{Multi-Scale Region-based Active Learning Pipeline.} In the proposed framework, the detector model is first trained on the initial labeled data $D_L$, and make the multi-scale inference on unlabeled data $D_U$ in different resolution, (a) Region candidates are generated from the multi-scale predictions using a greedy approach (Sec. \ref{subsec:region}). (b) Combine uncertainty-based scores and category re-weighting approach to evaluate the informative region score (Sec. \ref{subsec:IS}). (c) Then, a scale-aware selection method is used to select the diverse regions and allocate labeling budgets for each scale (Sec. \ref{subsec:RS}). (d) The selected regions are annotated from a human expert, and subsequently updated to the labeled set (Sec. \ref{subsec:acquisition}). The process repeats for a specified number of cycles.
   }
\label{fig:pipeline}
\end{figure*}

\section{Related Work}
\subsection{Object Detection with Label Efficiency}
In the recent years, object detection has become a popular research area, resulting in the development of several successful models such as Faster R-CNN (FRCNN) 
\cite{ren2015faster}, YOLO \cite{redmon2017yolo9000}, RetinaNet \cite{lin2017focal} , SSD \cite{liu2016ssd}. However, object detection still requires a large amount of time-consuming and labor-intensive annotation. Therefore, some label-efficient approaches such as semi-supervised learning and weakly supervised learning have been proposed to reduce the annotation cost while improving model performance.

Semi-supervised learning combines both labeled data and unlabeled data in a data-efficient manner to enhance its accuracy. Current research focuses primarily on two aspects: consistency-based and pseudo-labeling. 
Consistency-based semi-supervised methods \cite{tang2021proposal, jeong2019consistency} aim to leverage unlabeled data by using an original image and its perturbed image to reduce the discrepancy between their respective output predictions. Pseudo-labeling approaches \cite{sohn2020simple, li2020improving, wang2018towards, xu2021end, liu2021unbiased} use the model's predictions as temporary labels for unlabeled data to improve its performance.

Weakly-supervised learning is a label-efficient approach where the model is trained on only image-level labels, instead of fully annotated data, to minimize the annotation cost while still improving model performance \cite{ash2019deep, bilen2015weakly, song2014weakly, li2016weakly}. WSDDN \cite{bilen2016weakly} was the first to employ a CNN-based approach for selecting positive samples, combining the recognition and detection scores, which had a significant impact on subsequent works \cite{diba2017weakly,jie2017deep,tang2018pcl}.

\subsection{Active Learning}
Active learning selection strategy is a method used to select the most informative samples from an unlabeled data for manual annotation. Two common types of active learning selection strategies are uncertainty-based and diversity-based methods. 

Uncertainty-based methods select the samples that the model is most uncertain about, by using criteria such as softmax entropy, confidence or margin of network outputs\cite{wang2016cost,wang2014new}. \cite{gal2017deep,gal2016dropout} proposed Bayesian active learning with MC-Dropout to estimate model uncertainty. In contrast, diversity-based methods aim to select a diverse set of samples that cover different regions of the feature space \cite{ash2019deep, choi2021active,kirsch2019batchbald}. Both methods have different trade-offs and have been shown to be effective in different scenarios. \cite{yoo2019learning} proposed a loss prediction module to predict the unlabeled sample loss as a selection criterion. \cite{sinha2019variational} leveraged variational autoencoder to learn a latent space and an adversarial network to distinguish between unlabeled and labeled data.

For object detection, coarse-grained active learning strategies typically select entire images as the unit for annotation \cite{haussmann2020scalable, yuan2021multiple, wu2022entropy, kao2018localization, vo2022active, choi2021active, yu2022consistency}. Some works use classification scores as metrics to measure model uncertainty, as seen in works such as \cite{haussmann2020scalable,yuan2021multiple}. \cite{wu2022entropy} proposed an entropy-based Non-Maximum Suppression method that considers both instance-level uncertainty and diversity. The work of \cite{kao2018localization} introduced two metrics called ``localization tightness" and ``localization stability" to measure the informativeness of object localization. Furthermore, some researchers have focused on combining the uncertainty of localization with the uncertainty of classification. For example, \cite{choi2021active} proposed a mixture density network to aggregate these two sources of uncertainty, and \cite{yu2022consistency} considered the consistency of localization and predicted class distributions in augmented and unlabeled images. 

Fine-grained active learning strategies query partial labels within specific instances or regions, instead of selecting the entire image. For instance, \cite{wang2018towards} proposed a self-supervised sample mining technique that uses pseudo-labeling with high consistency samples and an active learning strategy to select low consistency samples within cross-image validation. \cite{desai2020towards} employed the coreset method \cite{sener2017active} and classification uncertainty to select instance labels. Meanwhile, \cite{laielli2021region} focused on querying informative bounding boxes by selecting diverse groups of objects.

In contrast to prior works, our proposed multi-scale region-based active learning approach can acquire informative regions that not only reduce label redundancy but also enable efficient training with multi-scale regions.


\section{Problem Statement}
\label{sec:problem}
The problem of active learning on object detection is associated with two data sets: a labeled set $\mathbf{D_{L}}$ and an unlabeled set $\mathbf{D_{U}}$, where usually $\lvert \mathbf{D_{L}} \rvert \ll \lvert \mathbf{D_{U}} \rvert$. Besides, the process of a single iteration can be summarized into the following steps: (1) Training the detection model $M_{d}$ with the current labeled set $\mathbf{D_{L}}$. (2) Using the trained $M_{d}$ to infer the unlabeled set $\mathbf{D_{U}}$ and record the inference results. (3) Selecting a subset $\Delta \mathbf{D_{U}} \subset \mathbf{D_{U}}$ that are most informative unlabeled samples by designed active learning method. (4) Requesting a human expert to annotate selected samples in  $\Delta \mathbf{D_{U}}$. (5) Updating $\mathbf{D_{L}}$ as its union with the subset, i.e., $\mathbf{D_{L}} = \mathbf{D_{L}} \cup \Delta \mathbf{D_{U}}$. Then, the entire process would execute $C$ iterations until the detection performance is converged. 

The \textit{problem objective} is to develop an effective sample-selecting strategy (i.e., active learning method) to let the most informative unlabeled samples be chosen and annotated. Then, the training with these newly labeled samples would assist the detection model in achieving enhanced performance with less annotation cost.


\section{Method}
Although active learning is a promising solution to address the costly annotation burden in object detection, existing methods still have limitations, as mentioned in Sec. \ref{sec:intro}. Hence, we propose {\ourFullname} (\ourAbbrname), which takes the strengths of both coarse-grained and fine-grained active learning methods, as we now present in detail.

\subsection{MuRAL Overview}
\label{subsec:mural}
As depicted in Fig. \ref{fig:pipeline}, MuRAL consists of three components: (1) Multi-scale Region Candidate Generation: regions with distinct perception fields are generated to make objects of various sizes in the image prominent (Sec. \ref{subsec:region}). (2) Informative Score Calculation: the informative score of region is computed by the weighted uncertainty of object detection results in its field (Sec. \ref{subsec:IS}). (3) Scale-aware Region Selection: Most informative region candidates in each scale are selected to be labeled (Sec. \ref{subsec:RS}). At last, we describe the details of region label acquisition in Sec. \ref{subsec:acquisition}.

Notably, MuRAL follows the process stated in Sec. \ref{sec:problem} with one different operation. Rather than updating the $\mathbf{D_{L}}$ with the selected unlabeled images, MuRAL annotates the selected region candidates (denoted as $\mathbf{D_{R}}$) and updates $\mathbf{D_{L}}$ by $\mathbf{D_{L}} = \mathbf{D_{L}} \cup \mathbf{D_{R}}$, enhancing the diversity of $\mathbf{D_{L}}$.


\subsection{Multi-scale Region Candidate Generation}
\label{subsec:region}
One troublesome challenge in object detection is the various size of objects. Previous fine-grained methods might still fail due to the fixed scale of perception field. To this end, we aim to generate region candidates with distinct perception fields, making the size-varied objects prominent.

Let $r \coloneqq (x, y, H, W)$ denote a region, where $(x, y)$ is the image coordinate of top-left corner of the region, and $H$, $W$ are the height and width, respectively.
As illustrated in Fig. \ref{fig:pipeline}(a), unlabeled images from $\mathbf{D_{U}}$ are first transformed to multiple scales inspired by \cite{singh2018sniper}. Then, the object detection model $M_d$ (trained on current $\mathbf{D_{L}}$) performs inference on each scaled unlabeled images. Next, a sliding window approach with a region $r$ that has the same size of original unlabeled images is applied to generate region candidates. We highlight that applying the fixed-size region to multi-scale images is crucial since the region candidates can be added as training samples straightforwardly, accelerating the training process.

The approach is demonstrated in Algorithm \ref{algo:candidate}. After generating initial region candidates $\mathbf{R_{init}}$ by applying the sliding window over the image (line 7-11), we apply a greedy strategy to select the region  candidate $r_{g}$ that contains the maximum number of predicted objects (line 13). Note that an object is  counted only if its bounding box is completely contained in the region. Next, giving  $r_{g}$, the object in other region candidates would be removed if it appearing in $r_{g}$. By this, we can avoid selecting region candidates that are too similar (frequently occurring objects) or too adjacent. At last, the process would repeat until all predicted objects in the image are enclosed in selected region candidates $\mathbf{R_{c}}$.

\begin{algorithm}[t]
\caption{Region Candidate Generation}
\label{algo:candidate}
\begin{algorithmic}[1]

\STATE {\bfseries Input:} image shape: ($H_{i}, W_{i}$)
\STATE {\bfseries Input:} region shape: ($H_{r}, W_{r}$)
\STATE {\bfseries Return:} region candidates: $\mathbf{R_{c}}$
\STATE
\STATE $\mathbf{R_{init}} \leftarrow \emptyset$
\STATE $\mathbf{R_{c}} \leftarrow \emptyset$
\FOR {x in ($0$, $H_{i} - H_{r}$)}
\FOR {y in ($0$, $W_{i} - W_{r}$)}
\STATE $\mathbf{R_{init}} \leftarrow \mathbf{R_{init}} \cup \{(x, y, H_{r}, W_{r})\}$
\ENDFOR
\ENDFOR
\WHILE {$\mathbf{R_{init}} \neq \emptyset$}
\STATE $r_{g} \leftarrow$  the region in $\mathbf{R_{init}}$ that contains most objects
\STATE $\mathbf{R_{c}} \leftarrow \mathbf{R_{c}} \cup \{r_{g}\}$
\FOR{ {\bfseries each} $r$ in $\mathbf{R_{init}}$}
\FOR{ {\bfseries each} predicted object $\hat{y_{o}}$ in $r_{g}$}
\STATE remove $\hat{y_{o}}$ in $r$ if it appears in $r$
\ENDFOR
\IF{$r$ contains no object}
\STATE $\mathbf{R_{init}} \leftarrow \mathbf{R_{init}} \setminus \{r\} $
\ENDIF
\ENDFOR
\ENDWHILE
\RETURN{} $\mathbf{R_{c}}$

\end{algorithmic}
\end{algorithm}


\subsection{Informative Score Calculation}
\label{subsec:IS}
After having the region candidates $\mathbf{R_{c}}$ from Sec. \ref{subsec:region}, we develop the informative score (IS) to assess the annotation priority of each region candidate. Intuitively, we should annotate the regions which cause the detection model $M_{d}$ to have an inferior detection result. Hence, the informative score aims to reflect the uncertainty of $M_{d}$ to the predicted objects in a region. Specifically, a single detection result of $M_{d}$ on an unlabeled image $I_{u}$ contains the predicted object category $\hat{y_{o}}$ (classification) and bounding box $\hat{y_{b}}$ (localization). And the object probability $\hat{y_{p}}$ (confidence) is computed by Eq. \ref{eq:confidence}, where $o \in \mathbf{O}$ is one of the object category, and $\mathbf{O}$ denotes all object categories.

\begin{equation}
    \hat{y_{p}} = P(\hat{y_{o}}=o \mid I_{u} ; M_{d})
    \label{eq:confidence}
\end{equation}

Assume a region candidate $r$ contains $K$ objects, and we collect their confidence scores as $\{\hat{y_{p}^{1}}, ... , \hat{y_{p}^{K}}\}$, where $\hat{y_{p}^{1}}$ is the confidence score of the first predicted object in the region, and so on. We calculate the informative score (IS) of $r$ by Eq. \ref{eq:informative}, which is the average uncertainty score to predicted objects in $r$. Intuitively, the region with higher uncertainty is more worth being annotated.
\begin{equation}
    IS(r) = \frac{1}{K} \sum_{i=1}^{K} \biggr[ 1-\hat{y^{i}_{p}} \biggl]
    \label{eq:informative}
\end{equation}

\begin{figure}
    \centering
    \includegraphics[width=\linewidth]{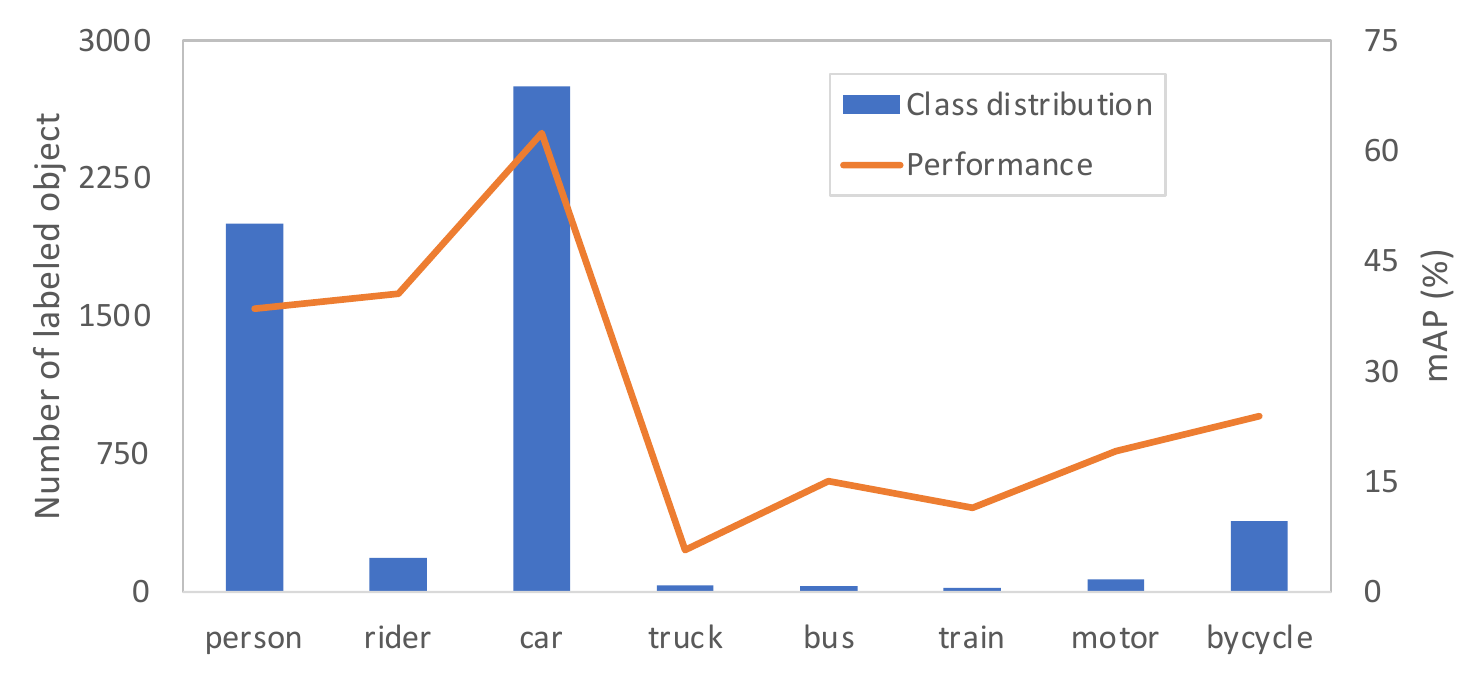}
    \caption{\textbf{Relation between class distribution and category performance on Cityscapes.} The findings suggest that the uneven distribution of classes can have an impact on the effectiveness of each category. This motivates us to design a class-balanced weighting module to alleviate this problem.}
    \label{fig:class}
    \vskip -0.3cm
\end{figure}

In addition, we observe that the category distribution of objects significantly affects the detection performance. Clearly, detection methods would over-focus on common objects and get poor performance on detecting rare objects, as shown in Fig. \ref{fig:class}. We should consider this factor when evaluating the informativeness of region candidates. Thus, we calculate the category distribution of objects using all ground-truth labels $\mathbf{Y_{o}}$ over images in $\mathbf{D_{L}}$ (Eq. \ref{eq:reweight}). Then, the reciprocals of probabilities are used to re-weight the informative score of a region. To be clear, the original uncertainty score is multiplied by the ground-truth probability $p(\hat{y^{i}_{o}})$ of $i$-th predicted object $\hat{y^{i}_{o}}$, as presented in Eq. \ref{eq:informative_final},

\begin{equation}
    p(o) = \frac{1}{\lvert \mathbf{Y_{o}} \rvert} \sum_{y_{o} \in \mathbf{Y_{o}}}  I(y_{o}=o)
    \label{eq:reweight}
\end{equation}
\begin{equation}
    IS_{final}(r) = \frac{1}{K} \sum_{i=1}^{K} \biggr[ \frac{1}{p(\hat{y_{o}^{i}})}(1-\hat{y^{i}_{p}}) \biggl]
    \label{eq:informative_final}
\end{equation}

where $I()$ is the indicator function, and $\hat{y^{i}_{p}}$ is the probability of $\hat{y^{i}_{o}}$. The aim of re-weighting is to encourage annotating region candidates that contain rare objects.

\subsection{Scale-aware Region Selection}
\label{subsec:RS}
After assigning the informative score, the annotation budget is allocated to these region candidates. Recall that we generate region candidates with multi-scale perception fields in Sec. \ref{subsec:region}. Let $\mathbf{R_{c}^{s}}$ denote the region candidates generated from the image scaled at scale $s$. Then, we have a list of multi-scale region candidates across scales $\{\mathbf{R_{c}^{1}}, ..., \mathbf{R_{c}^{s}}, ..., \mathbf{R_{c}^{T}}\}$. We emphasize that we have to be aware of the budget allocated to region candidates of different scales rather than directly choosing a region from their union to annotate (i.e., without the information of scales).

Algorithm \ref{alg:RS} elaborates on the progress of scale-aware region selection. Let $B$, $B_{acc}$, and $T$ denote the annotation budget in a training iteration, the accumulated annotation count, and the total scales, respectively. First, we sort all region candidates at different scales (line 5-7). Second, we iteratively loop over all $\mathbf{R_{c}^{s}}$ and annotate the region candidate with highest informative score (line 12-13). Third, we consistently collect newly annotated regions as $\mathbf{D_{R}}$ and terminate the process until $B_{acc}$ reaches $B$. We have promoted the diversity of selected samples via class distribution in Sec. \ref{subsec:IS}. From a different aspect, the region selection strategy is from various scales of perception fields, which is not taken into account in previous active learning studies. We carefully analyze its effectiveness in the experiment section.

\begin{algorithm}[t]
\caption{Scale-aware Region Selection}
\label{alg:RS}
\begin{algorithmic}[1]

\STATE {\bfseries Input:} annotation budget: $B$
\STATE {\bfseries Input:} region candidates across scales: $\{\mathbf{R^{1}_{c}}, ..., \mathbf{R^{T}_{c}} \}$
\STATE {\bfseries Return:} selected regions: $\mathbf{D_{R}}$
\STATE
\FOR{ {\bfseries each} $\mathbf{R^{s}_{c}}$ in $\{\mathbf{R^{1}_{c}}, ..., \mathbf{R^{T}_{c}} \}$}
\STATE Sort $\mathbf{R^{s}_{c}}$ by their informative scores (Eq. 4)
\ENDFOR
\STATE $\mathbf{D_{R}} \leftarrow \emptyset$
\STATE $B_{acc} \leftarrow 0$
\WHILE{$B_{acc} < B$}
\FOR{ {\bfseries each} $\mathbf{R^{s}_{c}}$ in $\{\mathbf{R^{1}_{c}}, ..., \mathbf{R^{T}_{c}} \}$}
\STATE $r \leftarrow $ pop the first region candidate from $\mathbf{R^{s}_{c}}$ 
\STATE Annotate all objects in $r$
\STATE $\mathbf{D_{R}} \leftarrow \mathbf{D_{R}} \cup \{r\}$
\STATE $n_{o} \leftarrow$ the number of annotated objects in $r$
\STATE $B_{acc} \leftarrow B_{acc} + n_{o}$
\IF{$B_{acc} \geq B$}
\STATE {\bfseries break}
\ENDIF
\ENDFOR
\ENDWHILE
\RETURN $\mathbf{D_{R}}$

\end{algorithmic}
\end{algorithm}
\vskip -0.2cm

\begin{figure*}
\begin{center}
\includegraphics[width=\linewidth]{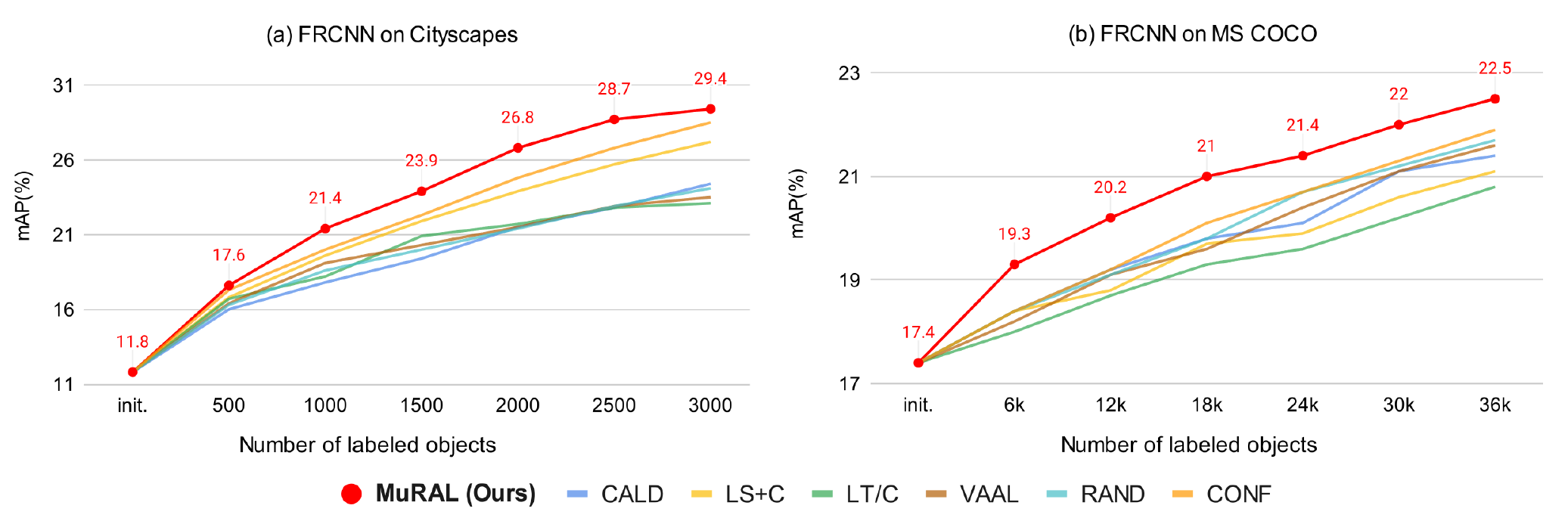}
\end{center}
   \vspace*{-4mm}
   \caption{\textbf{Comparison with different coarse-grained methods on 2 datasets.} 
   We conduct a comprehensive comparison of our region-based active selection strategy with other existing baselines and AL methods for object detection. For our experimental results on (a) Cityscapes \cite{cordts2016cityscapes} and (b) MS COCO \cite{lin2014microsoft}, demonstrate that \ourAbbrname{} outperforms all existing coarse-grained active selection approaches.
   }
\label{fig:coarse}
\end{figure*}

\begin{figure}
\begin{center}
\includegraphics[width=0.9\linewidth]{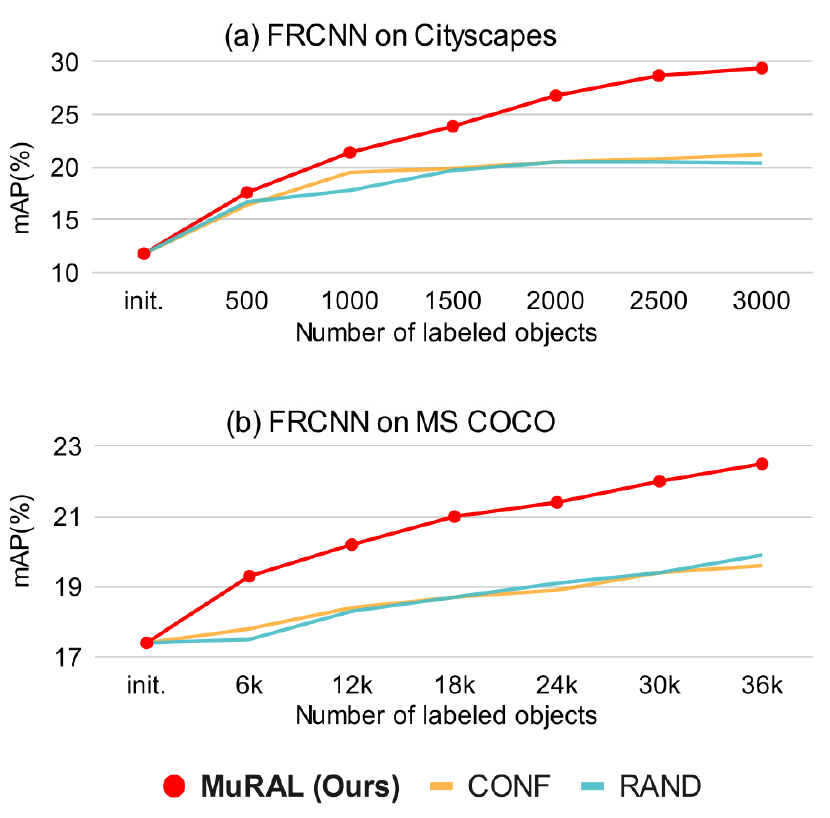}
\end{center}
   \vspace*{-1mm}
   \caption{\textbf{Comparison with fine-grained methods on 2 datasets.} 
   Our \ourAbbrname{} surpasses the fine-grained methods with random selection and uncertainty-based methods. It indicates that our approach is effective in improving training stability and mitigating the sampling bias described in Sec. \ref{subsubsec:fine}.
   }
\label{fig:fine}
\end{figure}

\subsection{Region Label Acquisition}
\label{subsec:acquisition}
After combining the final informative score and scale-aware region selection method, we obtained all selected regions $\mathbf{D_{R}}$. We then proceeded to annotate the objects within these regions $\mathbf{D_{R}}$ and update the labeled set $\mathbf{D_{L}}$, as described in Sec. \ref{subsec:mural}. For labeling, we considered all objects that partially overlapped (IoU $>$ 0.7) with a selected region, and we cropped all ground-truth boxes that partially overlapped with the region to use in the labeling process.

Additionally, we allocate the labeling budget as a fixed number of total ground truth bounding boxes instead of a fixed number of regions, as used in previous work \cite{desai2020towards}, because each region may contain a varying number of objects, and allocating the budget based on the number of regions could result in an uneven labeling effort during each active iteration. Therefore, we select bounding boxes in region candidates for annotation in the label acquisition process until the budget for the current round is fully utilized.

\section{Experiments}

\subsection{Experimental Settings}
\label{subsec:setting}

\noindent\textbf{Datasets.} 
To evaluate the effectiveness of MuRAL, we conduct experiments on two datasets, including Cityscapes \cite{cordts2016cityscapes} and MS COCO \cite{lin2014microsoft}. Cityscapes is a real-world autonomous driving dataset consisting of 2,975 training images, 500 validation images and 1,225 testing images with 8 categories, while MS COCO contains 118,287 training images and 5,000 validation images with 80 object categories.

As described in Sec. \ref{subsec:region}, we scale the original unlabeled image to multiple resolutions $(a, b)$, where $a$ represents the minimum size of the image and $b$ represents the maximum size. For instance, in MS COCO, we use the same multiple resolutions as described in \cite{singh2018sniper}, which are (480, 512), (800, 1280), and (1400, 2000). In Cityscapes, they are (2400, 3200), (1024, 2048), and (600, 1200). 

\vspace{1mm}

\noindent\textbf{Detector.} 
Our experiments used Faster RCNN \cite{ren2015faster} with a Resnet50 \cite{he2016deep} backbone as the detection model. Furthermore, we train for a total of 20 epochs in the initial training and each active iteration on 2 datasets.

\vspace{1mm}
\noindent\textbf{Implementation Details.} As stated before, we have a labeled set $\mathbf{D_{L}}$ and an unlabeled set $\mathbf{D_{U}}$ during training. In the experiment on Cityscapes, the initial size of $\mathbf{D_{L}}$ is only 1$\%$ (297 images) of all training images. Moreover, the annotation budget is set to 500 object labels(class, bounding box) in each iteration. Regarding the experiment on MS COCO, we use 5,000 images for the initial size of $\mathbf{D_{L}}$ followed by \cite{yu2022consistency} and then set the annotation budget to 6,000 object labels in each iteration. For all experiments, the active learning process contains 6 iterations, and the number of total scales $T$ is set to 3.

\newcommand{\tabincell}[2]{\begin{tabular}{@{}#1@{}}#2\end{tabular}} 
\begin{table*}[h]
\centering
\small
\setlength\tabcolsep{2.7pt}{
\begin{tabular}{c|ccc|cccccc}
\toprule 
\multicolumn{1}{c|}{} & \multicolumn{3}{c|}{Components} & \multicolumn{6}{c}{mAP wrt. labels} \\
\midrule
& \tabincell{c}{informative score} 
& \tabincell{c}{category weighting} 
&  \tabincell{c}{scale-aware}
& \tabincell{c}{500} & \tabincell{c}{1000} & \tabincell{c}{1500} & \tabincell{c}{2000} & \tabincell{c}{2500} & \tabincell{c}{3000}\\
\midrule
 & \multicolumn{3}{|c|}{CONF coarse-grained baseline} & 18.4 & 19.2 & 20.1 & 20.7 & 21.3 & 21.9\\
\midrule
(a) & \checkmark & &  & 16.2 & 18.2 & 19.8 & 21.2 & 21.6 & 21.8\\
(b) & \checkmark & \checkmark &  & 16.3 & 17.8 & 19.9 & 20.6 & 21.3 & 21.5\\
(c) & \checkmark& & \checkmark & 17.2 & 20.8 & 23.1 & 26.4 & 26.7 & 27.9\\
\cellcolor{LightCyan}(d) & \cellcolor{LightCyan}\checkmark & \cellcolor{LightCyan}\checkmark & \cellcolor{LightCyan}\checkmark & \cellcolor{LightCyan}\textbf{17.6} &  \cellcolor{LightCyan}\textbf{21.4} & \cellcolor{LightCyan}\textbf{23.9} & \cellcolor{LightCyan}\textbf{26.8} & \cellcolor{LightCyan}\textbf{28.7} & \cellcolor{LightCyan}\textbf{29.4}\\

\bottomrule  
\end{tabular}%
}
\vskip 0.2cm
\caption {\textbf{Ablation Studies on Cityscapes.} We compare the effectiveness of our proposed method, which includes the class re-weighting informative score and scale-aware selection method, with a coarse-grained uncertainty-based baseline method.}
\label{tab:Component-Ablation}%
\end{table*}%

\begin{table}[h]

\centering
\small
\setlength\tabcolsep{2.7pt}{
\begin{tabular}{c|ccc|cccccc}
\toprule 
\multicolumn{1}{c|}{} & \multicolumn{3}{c|}{Method} & \multicolumn{6}{c}{mAP wrt. labels} \\
\midrule
& \tabincell{c}{random} 
& \tabincell{c}{max} 
&  \tabincell{c}{mean}
& \tabincell{c}{500} & \tabincell{c}{1000} & \tabincell{c}{1500} & \tabincell{c}{2000} & \tabincell{c}{2500} & \tabincell{c}{3000}\\
\midrule
(a) & \checkmark & &  & 16.7 & 18.6 & 19.9 & 21.3 & 22.3 & 22.5\\
(b) &  & \checkmark &  &  \textbf{17.7} & 19.8 & 21.3 & 22.6 & 24.7 & 25.4\\ 
\cellcolor{LightCyan}(c) & \cellcolor{LightCyan} & \cellcolor{LightCyan} & \cellcolor{LightCyan}\checkmark & \cellcolor{LightCyan}{17.6} &  \cellcolor{LightCyan}\textbf{21.4} & \cellcolor{LightCyan}\textbf{23.9} & \cellcolor{LightCyan}\textbf{26.8} & \cellcolor{LightCyan}\textbf{28.7} & \cellcolor{LightCyan}\textbf{29.4}\\
\bottomrule  
\end{tabular}%
}
\vskip 0.2cm
\caption {\textbf{Comparison with different informative score.} Random refers to random score for candidate regions, Max uses the maximum score among each object covered in a region, and our proposed method computes the average confidence score (Eq. \ref{eq:informative_final}).}
\label{tab:region-ablation}%
\end{table}%


\subsection{Main Results}

We evaluate MuRAL with various active learning methods. For coarse-grained approaches, we include random selection, uncertainty-based selection \cite{wang2014new}, VAAL \cite{sinha2019variational}, CALD \cite{yu2022consistency}, LS+C and LT/C \cite{kao2018localization}. For fine-grained baselines, we utilize random selection and uncertainty-based selection, similar to \cite{desai2020towards}. Unfortunately, we are unable to compare with \cite{laielli2021region} since their source code is not available. Fig. \ref{fig:coarse} and Fig. \ref{fig:fine} demonstrate that our method surpasses all existing active learning methods across two datasets.

\subsubsection{Comparison with Coarse-grained Methods}
\label{subsubsec:coarse}
In the Citycapes dataset, the performance in Fig. \ref{fig:coarse} (a) shows that MuRAL outperforms all existing coarse-grained active learning baselines at each iteration, indicating its effectiveness in sampling informative and diverse regions. Moreover, MuRAL consistently outperforms all methods by more than 1 \% mean average precision (mAP) under all cases and also achieves an improvement of over 18\% within only 3000 newly labeled objects. Remarkably, we are able to achieve impressive results using annotation costs under 1\% of the total images (aroung 297 images) in each active learning cycle, suggesting that our methods effectively reduce the annotation cost. This is considering that an average image in Cityscapes contains around 18.5 objects, and annotating 1\% of the images is equivalent to labeling approximately 536 objects.

For MS COCO, our proposed method demonstrates superior performance compared to all other coarse-grained active learning methods, with at least a 0.4\% mAP improvement in each round and an overall 0.5\% mAP improvement over all methods, as illustrated in Fig. \ref{fig:coarse} (b). Impressively, even with just 36,000 annotations, which amounts to annotating only around 4\% (4931 images) of the dataset, considering the average number of objects in an image is approximately 7.3, our proposed method achieves state-of-the-art performance. These results demonstrate the effectiveness of our approach in minimizing annotation costs while achieving significant enhancements in object detection performance.

\begin{table*}[h]
\centering
\small
\setlength\tabcolsep{2.7pt}{
\begin{tabular}{c|c|c|ccccccccc}
\toprule 
dataset & method & \tabincell{c}{mAP} & \tabincell{c}{toaster} 
& \tabincell{c}{\footnotesize{hair}\\\footnotesize{drier}} & \tabincell{c}{scissors} & \tabincell{c}{ \footnotesize{parking}\\\footnotesize{meter}} & \tabincell{c}{toothbrush} & \tabincell{c}{snowboard} & \tabincell{c}{hotdog} & \tabincell{c}{refrigerator}\\
\midrule
\multirow{3}*{(a) MS COCO} & RAND & 21.7 & 2.6 & \textbf{0.7} & 5.3 & 27.4 & 2.4 & \textbf{11.2} & 11 & 27.3 \\
& CALD & 21.4 & \textbf{8.4} & 0 & 4.4 & 25.4 & 3.6 & 4.9 & 13.2 & 29.7 \\ 
& \cellcolor{LightCyan} Ours & \cellcolor{LightCyan}\textbf{22.5} & \cellcolor{LightCyan}{5.3} & \cellcolor{LightCyan}{0.3} & \cellcolor{LightCyan}\textbf{9.4} & \cellcolor{LightCyan}\textbf{30.2} & \cellcolor{LightCyan}\textbf{5.5} & \cellcolor{LightCyan}{10} & \cellcolor{LightCyan}\textbf{16.9} & \cellcolor{LightCyan}\textbf{32.7} \\
\toprule
dataset & method & \tabincell{c}{mAP} & \tabincell{c}{person} 
& \tabincell{c}{rider} & \tabincell{c}{car} & \tabincell{c}{truck} & \tabincell{c}{bus} & \tabincell{c}{train} & \tabincell{c}{motor} & \tabincell{c}{bike}\\
\midrule
\multirow{3}*{(b) Cityscapes} & RAND & 23.8 & 49.2 & 56.6 & \textbf{71.5} & 24.1 & 29.1 & 38.7 & 40 & 39.1 \\
& CALD & 24.6 & \textbf{52} & 60.3 & 69.9 & 26.5 & 26.2 & 27.9 & 42.8 & 41.5 \\ 
& \cellcolor{LightCyan}Ours & \cellcolor{LightCyan}\textbf{29.4} & \cellcolor{LightCyan}51 &  \cellcolor{LightCyan}\textbf{67.4} & \cellcolor{LightCyan}68.8 & \cellcolor{LightCyan}\textbf{26.6} & \cellcolor{LightCyan}\textbf{35.5} & \cellcolor{LightCyan}\textbf{49.3} & \cellcolor{LightCyan}\textbf{47.6} & \cellcolor{LightCyan}\textbf{46.2}\\
\bottomrule  
\end{tabular}%
}
\vskip 0.2cm
\caption {\textbf{Difficult Category Performance on (a) MS COCO and (b) Cityscapes.}
we conduct a comparative analysis using three distinct approaches. In MS COCO, we regard performance lower than 30\% AP as difficult classes. The results show that our \ourAbbrname{} are effective in improving the performance on difficult categories, outperforming other methods by selecting a higher number of challenging classes.
}
\label{tab:category}%
\end{table*}%

\begin{table}[h]
\centering
\small
\setlength\tabcolsep{2.7pt}{
\begin{tabular}{c|c|cccccc}
\toprule 
\multicolumn{1}{c|}{} & \multicolumn{1}{c|}{} & \multicolumn{6}{c}{mAP wrt. labels} \\
\midrule
& \tabincell{c}{Scale} 
& \tabincell{c}{500} & \tabincell{c}{1000} & \tabincell{c}{1500} & \tabincell{c}{2000} & \tabincell{c}{2500} & \tabincell{c}{3000}\\
\midrule
(a) & single scale & 17.5 & 19.9 & 21.3 & 22.6 & 23.1 & 23.3\\
(b) & 2 scale &  16.9 & 19.9 & 21.7 & 22.2 & 22.7 & 23.4\\ 
\cellcolor{LightCyan}(c) & \cellcolor{LightCyan}3 scale & \cellcolor{LightCyan}\textbf{17.6} &  \cellcolor{LightCyan}\textbf{21.4} & \cellcolor{LightCyan}\textbf{23.9} & \cellcolor{LightCyan}\textbf{26.8} & \cellcolor{LightCyan}\textbf{28.7} & \cellcolor{LightCyan}\textbf{29.4}\\
\bottomrule  
\end{tabular}%
}
\vskip 0.2cm
\caption {\textbf{The impact of multi-scale approach on Cityscapes.} 
Our results indicate that using three scales can lead to better performance compared to other configurations. Specifically, we found that using two scales with resolutions of (2400, 3200) and (1024, 2048) did not perform as well as the three-scale approach.}
\label{tab:scale}%
\vskip -0.3cm
\end{table}%

\subsubsection{Comparison with Fine-grained Methods}
\label{subsubsec:fine}
For fine-grained methods, we employ fine-grained methods that utilize random and uncertainty-based selection strategies to identify informative instances. Specifically, we sort the predicted bounding boxes according to either their random score or confidence score, and select a budget of bounding boxes for annotation. Moreover, during training, we ignore unlabeled instances in images that contain both labeled ground truth boxes and unlabeled bounding boxes.

As shown in Fig. \ref{fig:fine}, MuRAL outperforms the fine-grained methods by 8\% mAP on Cityscapes and 2.5\% mAP on MS COCO. In addition, the fine-grained methods demonstrate a marginal improvement in each active learning iteration, which can be attributed to the training instability caused by incomplete labels in images and sampling bias introduced by selecting non-diverse instances.

\subsection{Ablation Study}
\noindent \textbf{Componet Effectiveness} To evaluate the effectiveness of the proposed components, we conduct a detailed ablation study on Cityscapes. The results are presented in Tab. \ref{tab:Component-Ablation}. We highlight that the scale-aware region selection strategy mainly contributes to the performance, with an improvement of more than 8$\%$ mAP, which can be observed by comparing row (b) and (d). We attribute this phenomenon to the selected regions by the strategy containing higher diversity. The category weighting strategy (Eq. \ref{eq:informative} to Eq. \ref{eq:informative_final}) reveals its benefit (2$\%$ mAP improvement) to the performance when working with the scale-aware selection, as shown in row (c) and (d). These findings highlight the importance of considering different aspects of region diversity when selecting region candidates. Our components promote the diversity of selected regions from aspects of class distribution and multi-scale perception fields.

\noindent\textbf{Distinct Operations in Informative Score.} This ablation aims to verify the influence of applying different operations in informative score calculation: random, maximum, and average. The random operation randomly selects an uncertainty score as the informative score, while the maximum operation chooses the largest one. And the average operation (ours) computes the mean of all uncertainty scores (Eq. \ref{eq:informative_final}). As shown in Tab. \ref{tab:region-ablation}, the average operation achieves the best performance, improving mAP by over 4$\%$. The result suggests that the average operation can better represent the informativeness (uncertainty) of region candidates.

\vspace{2mm}

\noindent\textbf{Multiple Scales.} As mentioned earlier, we set the number of total scales as 3 for all experiments, which means the unlabeled images would be transformed into three different scales during the process of multi-scale region generation. This ablation attempts to verify the relationship between performance outcomes and different numbers of total scales, as presented in Table \ref{tab:scale}. It is evident that the performance consistently enhanced when the number of total scales increases. Comparing the results in the last column (newly added 3000 objects), the result of 3-scales outperforms the result of 1-scale by 6$\%$ mAP. The result underscores the importance of utilizing multi-scale samples during active learning training, which echoes our motivation.

\subsection{Case Study on Object Categories}
We develop MuRAL with multiple novel components and claim that MuRAL can detect those difficult-to-detect objects. This case study aims to validate the question, i.e., is MuRAL getting enhanced performance on object categories that are difficult to detect? To this end, we compare the performance of MuRAL with random selection and CALD (the state-of-the-art coarse-grained method) on the level of object category, as illustrated in Tab. \ref{tab:category}. Since MS COCO has many object categories, we treat and list categories with an AP lower than 30$\%$ as difficult classes. From the result in Tab. \ref{tab:category}(a), MuRAL outperforms baselines by over 4$\%$ mAP on difficult categories such as refrigerators and scissors. Similarly, from the results on Cityscapes (Tab. \ref{tab:category}(b)), MuRAL demonstrates improvements of 10$\%$ and 7$\%$ mAP on trains and bicycles (difficult categories), respectively.

\begin{figure}
    \centering
    \includegraphics[width=0.9\linewidth]{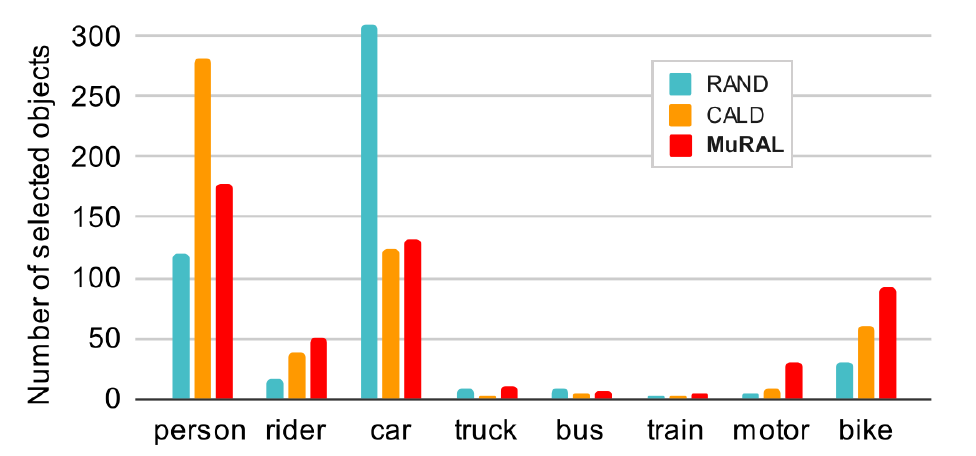}
    \vskip -1mm
    \caption{\textbf{Selected category distribution on Cityscapes.} We compare categorical distribution of selected data under different active selection strategies. The result indicate MuRAL is effective in addressing sampling bias by selecting more difficult categories and fewer instances from well-learned classes.}
    \label{fig:distribution}
    \vskip -0.4cm
\end{figure}

Furthermore, Fig. \ref{fig:distribution} highlights that MuRAL is particularly effective in selecting objects with lower performance while reducing the selection of easy-to-detect objects, resulting in a more balanced distribution of object categories.

\section{Conclusion}
We introduce a new approach for object detection called \ourFullname{} (\ourAbbrname{}), which addresses the issue of redundant labeling of well-learned objects and enhances training stability by selecting informative regions of multiple scales. Our approach considers informative scores based on class weighting to prioritize difficult-to-detect classes and employs a scale-aware selection strategy for sampling diverse regions. Experimental results demonstrate the effectiveness of our approach in outperforming existing active learning methods and improving the performance of difficult classes in object detection tasks.

{\small
\bibliographystyle{ieee_fullname}
\bibliography{egbib}
}

\clearpage
\newpage


\section*{Supplementary material}

# Multi-scale Region-based Active Learning for Object Detection
# Supplementary Material

Yi-Syuan Liou[1]   Tsung-Han Wu[1]   Jia-Fong Yeh[1]   Wen-Chin Chen[1]   Winston H. Hsu[1,2]

[1]National Taiwan University   [2]Mobile Drive Technology

In this supplementary material, we introduce more details on our implementation and existing active learning baseline methods. Moreover, we provide additional experimental results and visualization on MS COCO [4] and Cityscapes [1] with existing active learning methods.

## 1. Implementation Details

The overview of our **MuRAL** approach is presented in Sec. 4.1 of the main paper. For all experiments on Cityscapes and MS COCO, we train the model using a single 2080Ti GPU and set the batch size to 4. We employ Faster RCNN [5] with Resnet50 backbone [2] as the detection model. Furthermore, we utilize the SGD optimizer with an initial learning rate of 2.5e-3 and a decay rate of 0.1.

## 2. Active Learning Baselines

In this section, we provide more information about the baselines discussed in Sec. 5 of the main paper.

### 2.1. Coarse-grained Methods

For all experiments, we select all the informative samples from the unlabeled images based on the budget allocated for bounding boxes, following the settings explained in Sec. 4.5 of the main paper.
**RAND:** This method randomly selects images from the pool of unlabeled data for annotation.
**CONF:** Wang et al. [7] proposed an uncertain-based method for selecting unlabeled samples with the least confidence in their predictions. The confidence score of an unlabeled image is evaluated by averaging the softmax confidence values of all predicted objects as shown in Eq. 1.

$$S = \frac{1}{N} \sum_{i=1}^{N} P(\hat{y_o} = o \mid I_u; M_d) \qquad (1)$$

where the total number of instances and the detection model and unlabeled image are represented by N, $M_d$, and $I_u$, respectively.

**VAAL:** Sinha et al. [6] utilized a variational autoencoder to learn a latent space and an adversarial network to differentiate between unlabeled and labeled data. The approach selects unlabeled samples with uncertain representations.
**LS+C:** Kao et al. [3] proposed the localization stability metric, which evaluates the stability of each bounding boxes by unlabeled samples with noise.
**LT/C:** Kao et al. [3] introduced the localization tightness metric to evaluate unlabeled images with the consistency between the region proposal and the prediction. This metric measures the localization informativeness between true instances and detected bounding boxes.
**CALD:** Yu et al. [8] evaluated the localization and classification informative score, which considers the consistency between augmented and unlabeled images. They then utilized mutual information selection method to evaluate consistency of class distributions between labeled image and unlabeled image.

### 2.2. Fine-grained Methods

In fine-grained selection methods, we select informative instances from unlabeled data and annotate them based on the closest ground-truth object in the queried image. To avoid incomplete label training, we ignore all unlabeled instances in the queried image during the training process, as described in Sec. 5.2.2 of the main paper.
**RAND:** From all predicted instances in unlabeled images, we randomly select instances for annotation.
**CONF:** Following by [7], we select the least confident instances from the predicted unlabeled images.

## 3. Extensive Analyses and Results

### 3.1. Experimental Results

Due to space limitations, we include the original experimental results of Fig. 4 and Fig. 5 in the main paper in the Tab. 2 and 3. Furthermore, we present category performance comparisons with fine-grained methods in Tab. 1, and coarse-grained comparison results are provided in Sec. 5.4 of the main paper. The results show that our method

1

| dataset | method | mAP | toaster | hair drier | scissors | parking meter | toothbrush | snowboard | hotdog | refrigerator |
|---|---|---|---|---|---|---|---|---|---|---|
| MS COCO | RAND | 19.9 | 0 | 0 | 10.8 | 26.8 | 4.5 | 8.5 | 14.4 | 24.1 |
| | CONF | 19.6 | 0 | 0 | 8.2 | 24.7 | 3.6 | 8.5 | 15.4 | 22.2 |
| | **MuRAL** | **22.5** | **5.3** | **0.3** | **9.4** | **30.2** | **5.5** | **10** | **16.9** | **32.7** |

| dataset | method | mAP | person | rider | car | truck | bus | train | motor | bike |
|---|---|---|---|---|---|---|---|---|---|---|
| Cityscapes | RAND | 20.4 | 43.3 | 46.3 | 64 | 14 | 21.6 | 43.8 | 32.9 | 34.1 |
| | CONF | 21.2 | 44 | 47.1 | 63.5 | 15 | 21.3 | 41.8 | 34.6 | 35.3 |
| | **MuRAL** | **29.4** | **51** | **67.4** | **68.8** | **26.6** | **35.5** | **49.3** | **47.6** | **46.2** |

Table 1. **Category Performance Comparison with fine-grained methods on MS COCO and Cityscapes.** Experimental result shows that our method outperforms fine-grained methods on each category performance. Note that we list difficult categories with an AP lower than 30% in MS COCO as described in Sec. 5.4 of main paper.

| | method | | mAP wrt. labels | | | | | |
|---|---|---|---|---|---|---|---|---|
| | | init. | 0.5k | 1k | 1.5k | 2k | 2.5k | 3k |
| Coarse-grained | CALD | 11.8 | 16 | 17.8 | 19.4 | 21.5 | 22.8 | 24.4 |
| | LS+C | 11.8 | 16.8 | 19.6 | 21.9 | 23.9 | 25.7 | 27.2 |
| | LT/C | 11.8 | 16.7 | 18.2 | 20.9 | 21.7 | 22.8 | 23.1 |
| | VAAL | 11.8 | 16.4 | 19.1 | 20.3 | 21.5 | 22.9 | 23.5 |
| | RAND | 11.8 | 16.3 | 18.6 | 20 | 21.4 | 22.9 | 24.1 |
| | CONF | 11.8 | 17.3 | 20 | 22.3 | 24.8 | 26.8 | 28.5 |
| Fine-grained | RAND | 11.8 | 16.7 | 17.8 | 19.7 | 20.5 | 20.5 | 20.4 |
| | CONF | 11.8 | 16.4 | 19.5 | 19.9 | 20.5 | 20.8 | 21.2 |
| **MuRAL** | | 11.8 | **17.6** | **21.4** | **23.9** | **26.8** | **28.7** | **29.4** |

Table 2. Detailed performance of all methods evaluated on Cityscape. Mean Average Precision (mAP) is reported. The proposed MuRAL demonstrates the superior performance in each active learning iteration.

| | method | | mAP wrt. labels | | | | | |
|---|---|---|---|---|---|---|---|---|
| | | init. | 6k | 12k | 18k | 24k | 30k | 36k |
| Coarse-grained | CALD | 17.4 | 18.4 | 19.2 | 19.8 | 20.1 | 21.1 | 21.4 |
| | LS+C | 17.4 | 18.4 | 18.8 | 19.7 | 19.9 | 20.6 | 21.1 |
| | LT/C | 17.4 | 18 | 18.7 | 19.3 | 19.6 | 20.2 | 20.8 |
| | VAAL | 17.4 | 18.2 | 19.1 | 19.6 | 20.4 | 21.1 | 21.6 |
| | RAND | 17.4 | 18.4 | 19.1 | 19.8 | 20.7 | 21.2 | 21.7 |
| | CONF | 17.4 | 18.4 | 19.2 | 20.1 | 20.7 | 21.3 | 21.9 |
| Fine-grained | RAND | 17.4 | 17.5 | 18.3 | 18.7 | 19.1 | 19.4 | 19.9 |
| | CONF | 17.4 | 17.8 | 18.4 | 18.7 | 18.9 | 19.4 | 19.6 |
| **MuRAL** | | 17.4 | **19.3** | **20.2** | **21** | **21.4** | **22** | **22.5** |

Table 3. Detailed performance of all methods evaluated on MS COCO. Mean Average Precision (mAP) is reported. The proposed MuRAL demonstrates the superior performance in each active learning iteration.

outperforms both coarse-grained and fine-grained methods, especially in difficult category performance. It is worth noting that some difficult-to-detect classes, such as toaster and hair drier, have relatively low proportions within the dataset, which may have impacted their performance in MS COCO.

## 3.2. Visualization

**Multi-scale Regions.** Fig. 1 illustrates the visualization process of generating multi-scale regions. Initially, region candidates are generated using the greedy method outlined in Section 4.2 of the main paper (Fig. 1(a)). Next, we select informative and diverse regions based on their informative score and scale-aware selection, as depicted in the Fig. 1(b).

**Visualization Results.** Fig. 2 presents additional results for qualitative comparisons with ground truth and coarse-grained methods (RAND and CALD [8]). The results, particularly the area circled in red in Figure 2, demonstrate that our proposed **MuRAL** performs well on challenging categories such as bus and motor.

## 4. Limitations and Future Work

In this paper, we propose a multi-scale region-based active learning method to address the issue of label redundancy in coarse-grained methods and the problems of training instability and sampling bias in fine-grained methods. Our proposed approach effectively reduces the annotation cost and ensures the sampling of diverse regions to enhance object detection performance. However, there are still some areas that require further investigation. For example, generating region candidates may take more time for using more scales. The effectiveness of class re-weighting informative scores may still be restricted by the limited dataset size, particularly for some categories with few examples in MS COCO. Additionally, the class re-weighting approach may result in a trade-off between the performance of well-learned categories and difficult classes.

In light of the aforementioned areas for improvement, future research could explore ways to enhance the efficiency of generating region candidates. Additionally, it may be worthwhile to investigate the development of informative scores that consider both classification and localization aspects in region-based active learning for object detection.

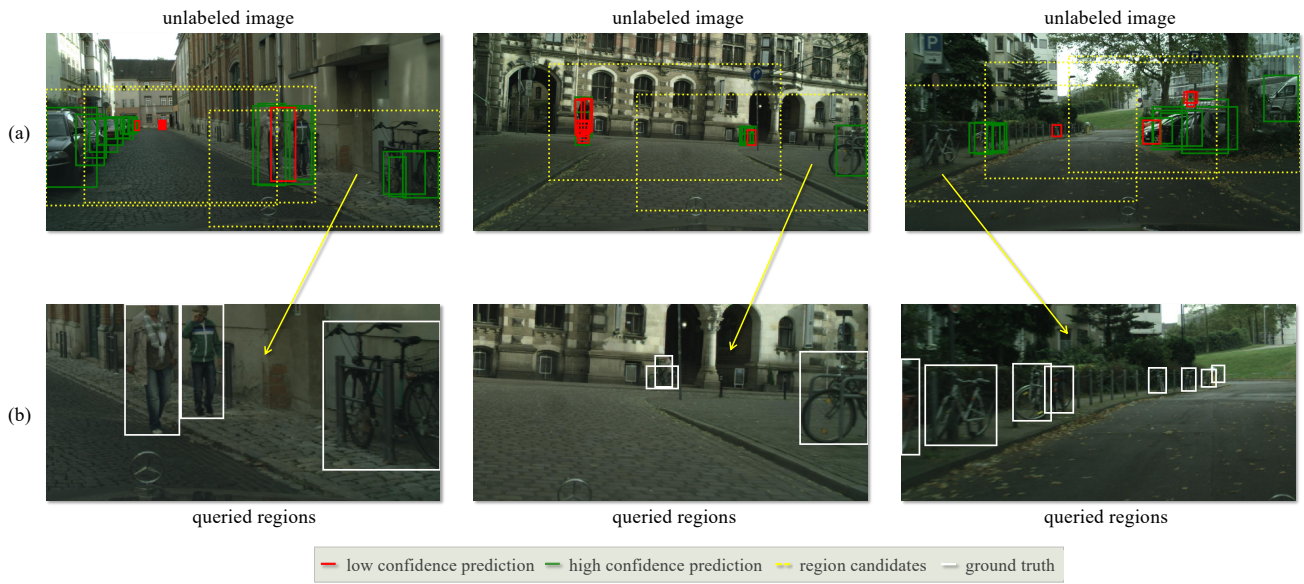

Figure 1. **Visualization of multi-scale region candidate generation and selected informative regions on Cityscapes.**

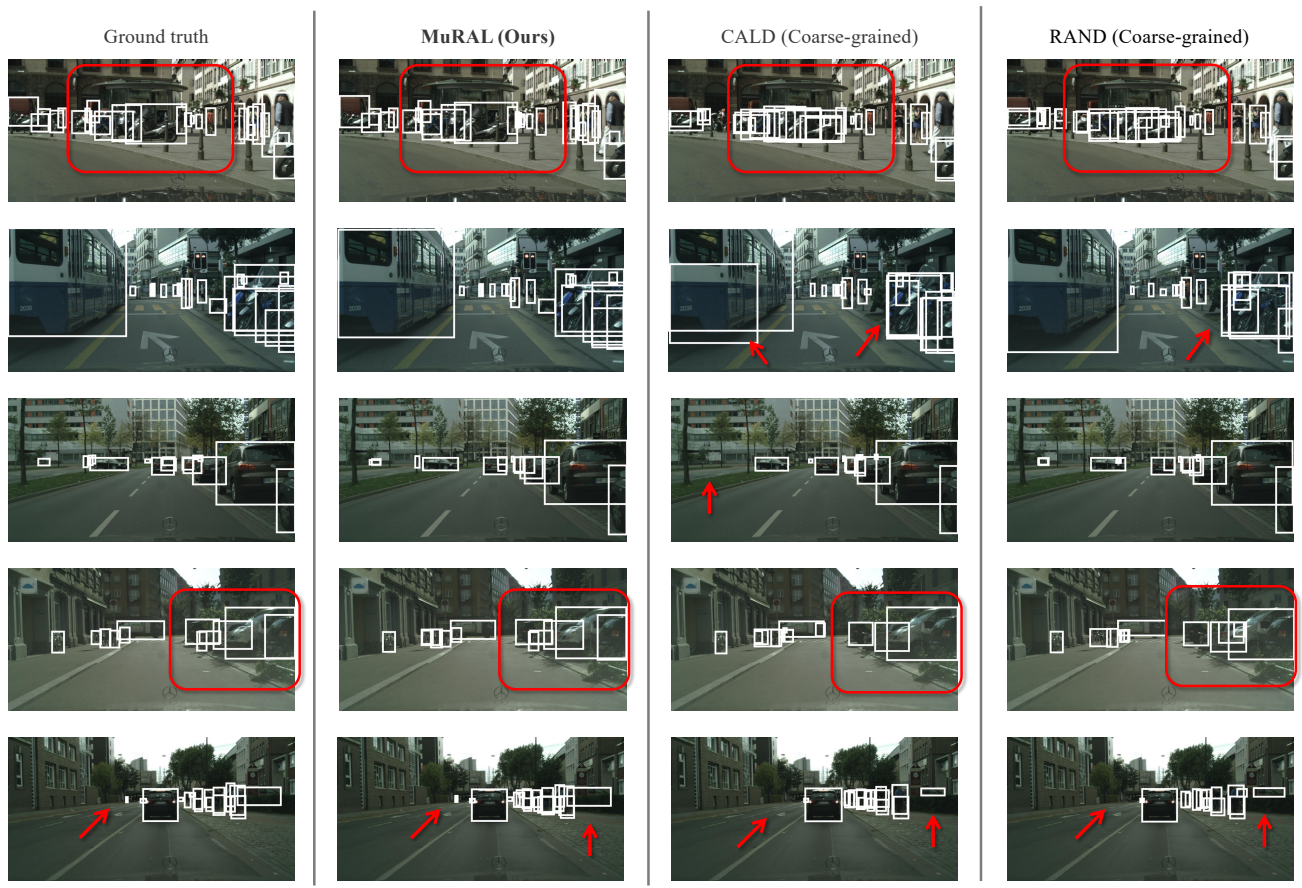

Figure 2. **Visualization results compared with ground truth and coarse-grained methods (RAND & CALD) on Cityscapes.** The highlighted areas in the figure, as indicated by the red arrows and circles, demonstrate that our proposed MuRAL performs well on challenging objects and achieves accurate predictions.



\end{document}

%% file: custom.tex
\usepackage[nocompress]{cite}
\usepackage{multirow}
\usepackage{xcolor}
\usepackage{setspace}
\usepackage{amsfonts}
\usepackage{colortbl}
\definecolor{gray}{rgb}{0.5,0.5,0.5} 
\definecolor{frenchblue}{rgb}{0.0, 0.45, 0.73}
\definecolor{gray}{rgb}{0.5,0.5,0.5} 
\definecolor{green}{rgb}{0, 0.4, 0} 
\definecolor{orange}{rgb}{1, 0.5, 0} 	
\definecolor{mahogany}{rgb}{0.75, 0.25, 0.0}
\definecolor{purple}{rgb}{0.6, 0, 0.6}
\definecolor{darkgreen}{rgb}{0, 0.4, 0.4} 
\definecolor{red}{rgb}{1.0, 0, 0}
\definecolor{plotpurple}{rgb}{0.2353, 0.2, 0.90196}
\definecolor{plotorange}{rgb}{1.0, 0.6, 0.2}
\definecolor{plotgreen}{rgb}{0.2, 0.784313, 0.2}
\definecolor{plotred}{rgb}{1.0, 0.2, 0.392}
\definecolor{LightCyan}{rgb}{0.88,1,1}

